\documentclass[letterpaper, 10 pt, conference]{ieeeconf}  % Comment this line out if you need a4paper

\IEEEoverridecommandlockouts                              % This command is only needed if 
\usepackage{amsmath} % assumes amsmath package installed
\usepackage{amssymb}  % assumes amsmath package installed
\usepackage{graphicx}
\usepackage{caption}
\usepackage{algorithm}
\usepackage{algorithmic}
\usepackage{subcaption}
\usepackage{array}
\usepackage{ragged2e}

\title{\LARGE \bf
State-of-the-art in Robot Learning for Multi-Robot Collaboration: A Comprehensive Survey
}

\author{Bin Wu and C. Steve Suh% <-this % stops a space
\thanks{The authors are with the Department of Mechanical Engineering, 
        Texas A\&M University, College Station, TX, 77840, USA
        {\tt\small $\{$wubin, ssuh$\}$@tamu.edu}}%
}

\begin{document}

\maketitle
\thispagestyle{empty}
\pagestyle{empty}

%%%%%%%%%%%%%%%%%%%%%%%%%%%%%%%%%%%%%%%%%%%%%%%%%%%%%%%%%%%%%%%%%%%%%%%%%%%%%%%%
\begin{abstract}
    With the continuous breakthroughs in core technology, the dawn of large-scale integration of robotic systems into daily human life is on the horizon. Multi-robot systems (MRS) built on this foundation are undergoing drastic evolution. The fusion of artificial intelligence technology with robot hardware is seeing broad application possibilities for MRS. This article surveys the state-of-the-art of robot learning in the context of Multi-Robot Cooperation (MRC) of recent. Commonly adopted robot learning methods (or frameworks) that are inspired by humans and animals are reviewed and their advantages and disadvantages are discussed along with the associated technical challenges. The potential trends of robot learning and MRS integration exploiting the merging of these methods with real-world applications is also discussed at length. Specifically statistical methods are used to quantitatively corroborate the ideas elaborated in the article.
\end{abstract}


%%%%%%%%%%%%%%%%%%%%%%%%%%%%%%%%%%%%%%%%%%%%%%%%%%%%%%%%%%%%%%%%%%%%%%%%%%%%%%%%
\section{INTRODUCTION}
%研究背景：介绍多机器人系统的重要性和应用领域
%问题陈述：阐述多机器人合作中存在的挑战。
%机器人学习的作用：简要说明如何通过机器人学习解决这些挑战。
%论文目的和贡献：概述论文的主要目标和预期贡献。
With the advancement of technology and the development of  artificial intelligence \cite{sze2017hardware, peddie2023history, tyagi2022machine}, robot learning has become one of the key factors driving the progress of robotic technology. Especially in the field of MRS \cite{arai2002advances}, robot learning has shown great potential and application value. MRS complete complex tasks by coordinating the actions of multiple robots, offering higher efficiency, reliability, and flexibility compared to single robot systems (SRS) \cite{rizk2019cooperative,fierro2018multi}. However, with the diversification of application scenarios and the increase in task requirements, the learning mechanisms in MRS face many challenges, including collaborative learning, communication constraint, environmental adaptability, and algorithm's ability to generalize \cite{kapoor2018multi, rogers2012delivering}.

This article reviews the latest research in robot learning within MRS, including theoretical foundations, key technologies, applications, challenges faced and paths being explored. An extensive review of existing literature allows the core issues and proper solution strategies to be identified along with the performance and limitations of different learning methods for practical applications.

First, the basic concepts of MRS and robot learning are introduced and their importance in current technological context is stated. Next, the design principles of learning mechanisms in MRS is explored in-depth, including, but not limited to, technologies such as Reinforcement Learning (RL), Transfer Learning (TL), and Imitation Learning (IL). Much insight is gained from evaluating the advantages and disadvantages of different learning strategies essential to inspiring future research and charting the path moving forward.  Moreover, successful case studies of MRS in specific applications are examined, such as automated warehousing, search and rescue, environmental monitoring, and precision agriculture, to showcase the effectiveness and applicability of robot learning technologies in solving real-world problems. The main challenges encountered in implementing multi-robot learning systems, including resource allocation, task decomposition, learning efficiency, and system scalability are also discussed.  
Finally, the article looks forward to the various directions of future development of robot learning in MRS, emphasizing the importance of interdisciplinary collaboration and how the integration of artificial intelligence, machine learning, control theory, and cognitive science will elevate the field to a greater level of development. The survey paper should provide researchers in the related fields with a comprehensive reference framework, inspire more innovative research and application, and jointly promote the progress of MRS and robot learning.

\section{DEFINITION AND SCOPE}
In this section, more detailed definitions of the key concepts addressed in this article are provided along with delineation of the scope of current issues. This is done to help establish a unified cognitive basis and also facilitate a deeper discussion of the issues based on the basis.
%多机器人系统， 多机器人合作，机器人学习，问题定义

\subsection{Multi-robot Systems}
\begin{figure}[htbp]
\vspace{4pt}
\centerline{\includegraphics[width=0.4\textwidth]{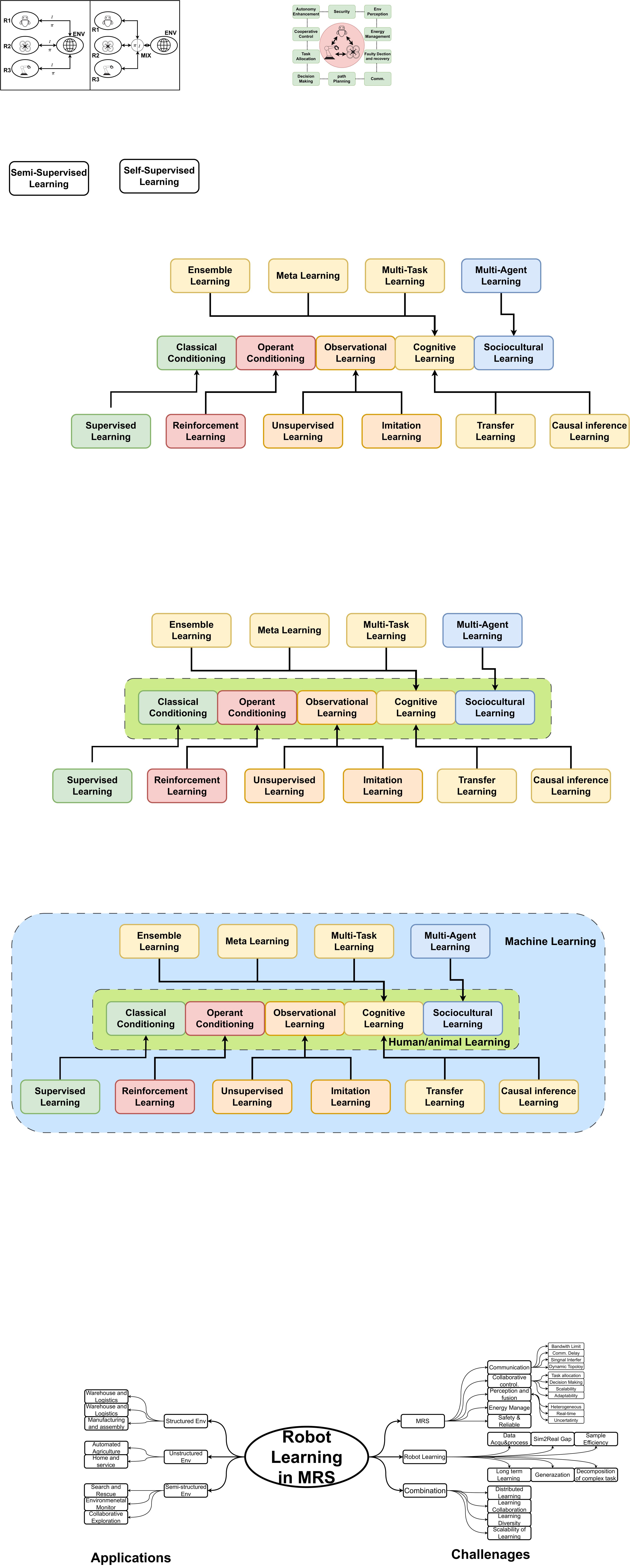}}
\captionsetup{font=footnotesize}

\caption{Single robot vs multi-robot interaction decision framework}
\label{fig: single vs multi}
\end{figure}

From a system-level perspective, the advantages of MRS over SRS are evident \cite{arai2002advances, gautam2012review}.
Figure \ref{fig: single vs multi} indicates the fundamental differences of the two systems.  An MRS refers to a system composed of two or more robots that can complete specific tasks or objectives through collaboration or competition. In such systems, each robot may have unique capabilities and limitations, and they achieve collective intelligence and action through communication and coordination. Research on MRS mainly focuses on how to effectively design and implement interaction, communication, and collaboration mechanisms among robots to improve the efficiency and effectiveness of the entire system \cite{gautam2012review}.

An MRS can be represented as a tuple $(N, S, A, T, R, C, G)$.

\begin{itemize}
    \item Robot Set: Let $N = \{1, 2, \ldots, n\}$ be the set of robots in an MRS, where each $i$ represents an independent robot entity. 
    \item State Space: For each robot $i$, its state can be represented by an element in the state space $S_i$. The state space of the entire system is the Cartesian product of these individual state spaces, i.e., $S = S_1 \times S_2 \times \ldots \times S_n$. 
    \item Action Space: Each robot $i$ can perform a series of actions, defined by the action space $A_i$. Similarly, the action space of the entire system is the Cartesian product of the individual action spaces, i.e., $A = A_1 \times A_2 \times \ldots \times A_n$. 
    \item Transition Function: The transition function $T: S \times A \rightarrow S$ defines how the system state changes based on the combination of actions performed. For a given current state and combination of actions, the transition function returns the next state of the system. 
    \item Reward Function: $R: S \times A \rightarrow \mathbb{R}$ assigns a real number reward value to each state and action combination, reflecting the effectiveness of that combination in achieving the system's goals. 
    \item Communication Model: In an MRS, communication between robots can be represented by the communication model $C$, which defines how robots exchange information with other robots or the system. 
    \item Goal Function: MRS usually have one or more goals, which can be represented by the goal function $G: S \rightarrow \mathbb{R}$, evaluating the extent to which the system achieves its goals in a specific state.
\end{itemize}

\subsection{Multi-robot Cooperation}
\begin{figure}[htbp]
\vspace{4pt}
\centerline{\includegraphics[width=0.4\textwidth]{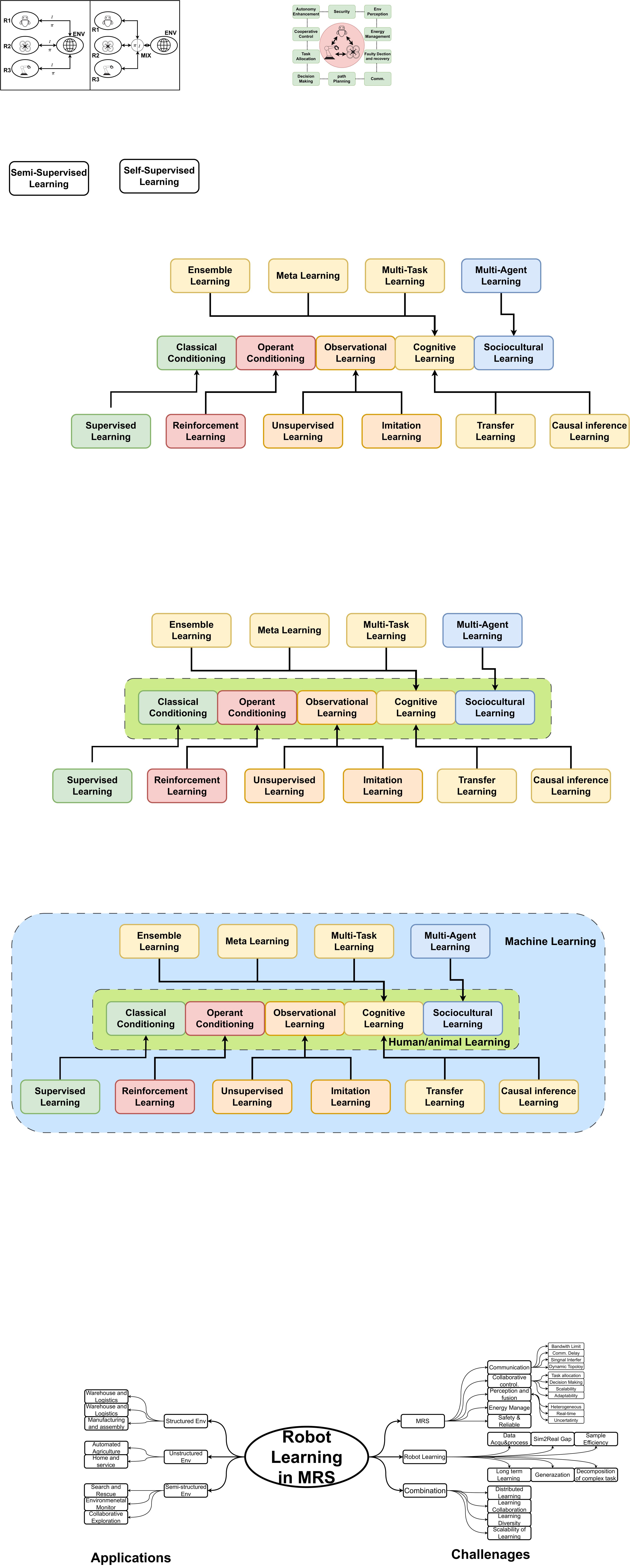}}
\captionsetup{font=footnotesize}

\caption{Sub-problems of multi-robot systems}
\label{fig: multi-robot tasks}
\end{figure}

\begin{figure*}[htbp]
\centering
\vspace*{5pt}
\includegraphics[width=1.0\textwidth]{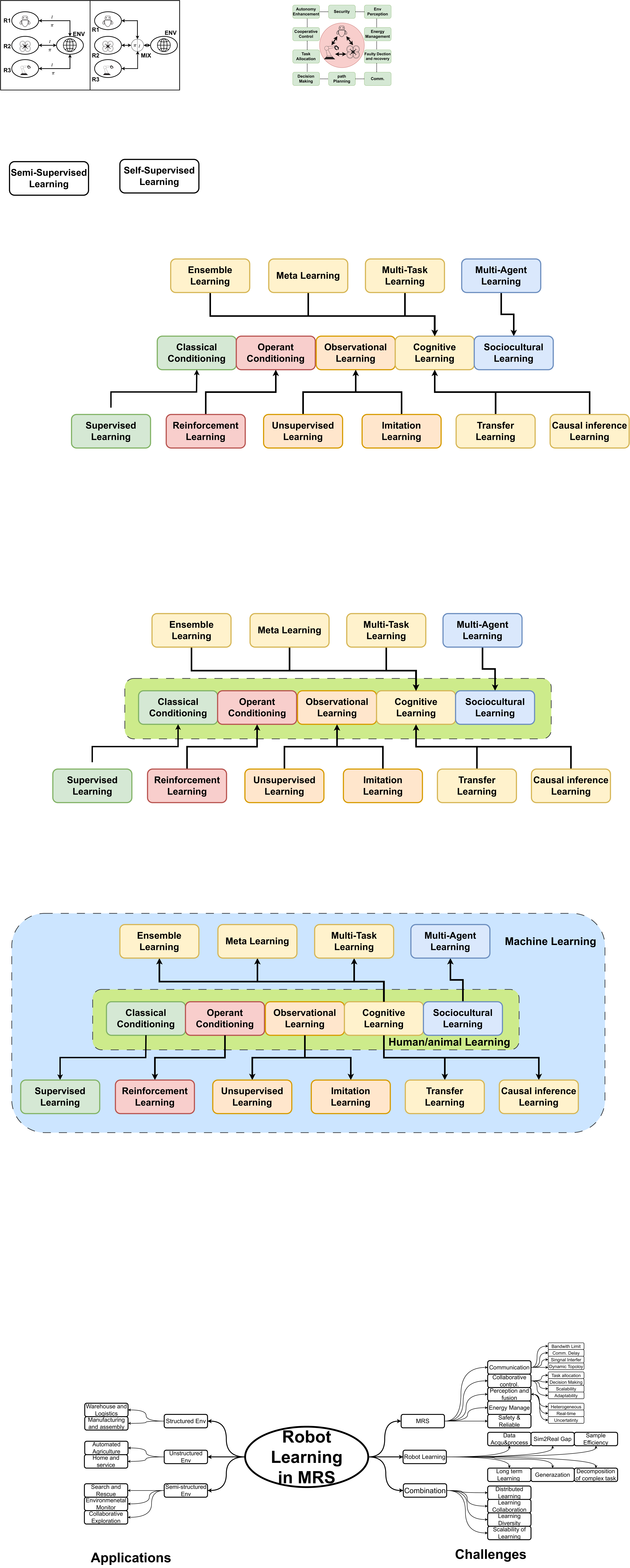}
\caption{Robot learning in the context of MRS: applications and challenges}
\label{fig:application and challenges}
\end{figure*}

The concept and mathematical definition of MRC problems can be further developed based on the foundational framework of MRS in the last section \cite{rizk2019cooperative,fierro2018multi}. MRC problems as seen in Figure \ref{fig: multi-robot tasks} involve a group of robots that share information, coordinate actions, and make decisions together to achieve a common task or goal. A more detailed application is provided by Figure \ref {fig:application and challenges}. The primary objective of cooperation is to utilize the collective capabilities of multiple robots to accomplish tasks that are not possible or inefficient for a single robot. This cooperation may include aspects such as task allocation, joint decision-making, resource sharing, coordinated actions, and goal sharing. To achieve this, a common goal function $( G: S \rightarrow \mathbb{R} )$ can be defined to measure and observe the performance of the entire system in achieving the common goal.

\begin{itemize}
    \item Collaboration Strategy: Define a set of strategies $ \Pi = \{\pi_1, \pi_2, \ldots, \pi_n\} $, where each $ \pi_i: S \rightarrow A_i $ is a decision rule guiding robot $i$ in choosing actions in a given state. 
    \item Communication Model Extension: Extend the communication model $C$ to $Ce: S \times N \rightarrow \mathcal{P}(N)$, where $ \mathcal{P}(N) $ is the power set of the robot set $\boldsymbol{N} $, indicating which groups of robots can communicate in a given state.
    \item Joint Action: Define a joint action mapping $\alpha: S \times \Pi \rightarrow A $, combining the strategies of all the robots and current state to produce the action of the entire system.
    \item Collaborative Benefit: Introduce a utility function $U: S \times A \rightarrow \mathbb{R}$ to evaluate the performance of the system as a whole under a given state and action.
    \item Constraints: Consider the constraints of interactions and cooperation among robots, such as communication range, resource limitation, and task dependency.
\end{itemize}

\subsection{Robot Learning}
When discussing robot learning, researchers often first consider it as an interdisciplinary field bringing together machine learning and robotics, which is undoubtedly correct. However, before clarifying the essence of robot learning, what learning is must be clarified. In sociology \cite{berger1967social}, learning is usually defined as the process by which individuals or groups acquire social behavior patterns through the process of socialization. This includes the internalization of social norms, values, language, skills, and behavior patterns. Sociologists emphasize the impact of social structures and culture on the individual learning process. 
The definition of learning in anthropology \cite{lave1991situated} emphasizes the process of cultural transmission and adaptation. Learning is seen as the way individuals acquire knowledge, skills, beliefs, and behavior patterns from their cultural environment. Anthropologists often focus on the differences in learning across cultures or ethnic groups and how these differences affect the adaptation and development of individuals and communities. 
In psychology \cite{bandura1977social}, learning is typically defined as a relatively permanent change in behavior or thought patterns produced by experience. This definition emphasizes an individual's response and adaptation to environmental stimuli, including cognitive learning, emotional learning, and behavioral learning.

In traditional disciplines, there are far more definitions of learning than those listed above, but this does not prevent one from finding some commonalities. In this article, the concept of learning from four dimensions is discussed: 
\begin{itemize}
    \item Utilization of Experience: Learning is a process of acquiring knowledge or skills through experience.
    \item Behavioral and Cognitive Changes: Learning leads to changes in behavior or ways of thinking.
    \item Environmental Adaptation: Learning is considered a way for individuals or groups to adapt to their environment.
    \item Process Nature: Learning is viewed as a continuous process, rather than a single event. This process involves the constant acquisition, processing, and application of information, as well as gradual adaptation and change over time.
\end{itemize}

Returning to the definition of robot learning, it refers to the process where a robot merges hardware (such as sensing, processing, and executing components) with software (specific learning algorithms) to utilize data, experience, or interactions with the environment to acquire new knowledge or skills, thereby improving its performance \cite{kroemer2021review,connell2012robot,ravichandar2020recent}.
This process involves perception (acquiring information through sensors), decision-making (making decisions based on learned knowledge), and action (executing tasks). The goal of robot learning is to enable robots to autonomously adapt to new tasks and environments, enhancing their flexibility and efficiency. Suppose a robot's strategy $( \pi: S \rightarrow A )$ is a mapping from state $s$ to action $a$, guiding the robot in choosing actions in a given state, then, the learning algorithm is used to extract patterns and knowledge from the data to improve the strategy $( \pi )$.

%定义和分类：定义多机器人系统，描述其主要类型和分类。
%应用场景：探讨多机器人系统在不同领域（如搜索与救援、工业自动化等）的应用。
%机器人学习概述：介绍机器人学习的基本原理。
%常用算法：评述用于机器人学习的主要算法，如强化学习、监督学习等。
%学习挑战：讨论在多机器人环境中实施机器人学习所面临的特殊挑战。
\subsection{Robot Learning in MRC}
In the section above, the concepts of MRC and robot learning were separately introduced. Building on this foundation, it naturally leads to another question: what is the difference between robot learning in MRC and single robot learning?  Before answering this question, an assumption needs to be made that, regardless if MRS or SRS is being considered, each robot makes decisions based on the local information it acquires. Below, the question is addressed from two aspects: 1. Utilization of information, where robots in a MRS exchange information. 2. Joint decision-making, whether it's competitive or cooperative decision-making, there is a coordination mechanism among the robots in a MRS to promote more efficient operation of the overall system. These two basic points are both the advantage and challenge of multi-robot learning. Exchange of information (experience) provides the robots with more learning materials but also increases learning burden. Joint decision-making can create a synergy effect, while also imposing more restrictions on each robot's decision-making, increasing the difficulty of learning.

\section{Learning Method}
The concept of machine learning was inspired by the ways human and animal learn \cite{rosenblatt1958perceptron, mcculloch1943logical}. Researchers design algorithms and machines by simulating how the human brain works, hoping that machines can acquire knowledge and skill through observation and experience. The classification of learning methods discussed in this section is also based on a logical division, mirroring human or animal learning methods. Afterward various robot learning methods is discussed in the context of MRS.

\subsection{Carbon-based vs. Silicon-based}
First, one can draw from psychology and neuroscience the following classifications of human and animal learning methods:
\begin{itemize}
    \item Classical Conditioning: Conducted by the Russian physiologist Pavlov \cite{pavlov2010conditioned}, who discovered how animals learn through associating stimuli with responses via experiments.
    \item Operant Conditioning: Introduced by B.F. Skinner in his book \cite{skinner2019behavior}, the concept involves increasing or decreasing the frequency of specific behaviors through rewards and punishments.
    \item Observational Learning: Individuals learn by observing others' behaviors and their consequences \cite{bandura1977social}.
    \item Cognitive Learning: Emphasizes the role of exploration and problem-solving in the learning process \cite{piaget1952origins}.
    \item Sociocultural Learning: Discusses how social interactions influence cognitive development \cite{lev1979mind}.
    \item Affective Learning: Explores how emotions affect learning and memory, especially the neural mechanisms of fear responses \cite{ledoux1998emotional}.
\end{itemize}

\begin{figure*}[htbp]
\centering
\vspace*{5pt}
\includegraphics[width=1.0\textwidth]{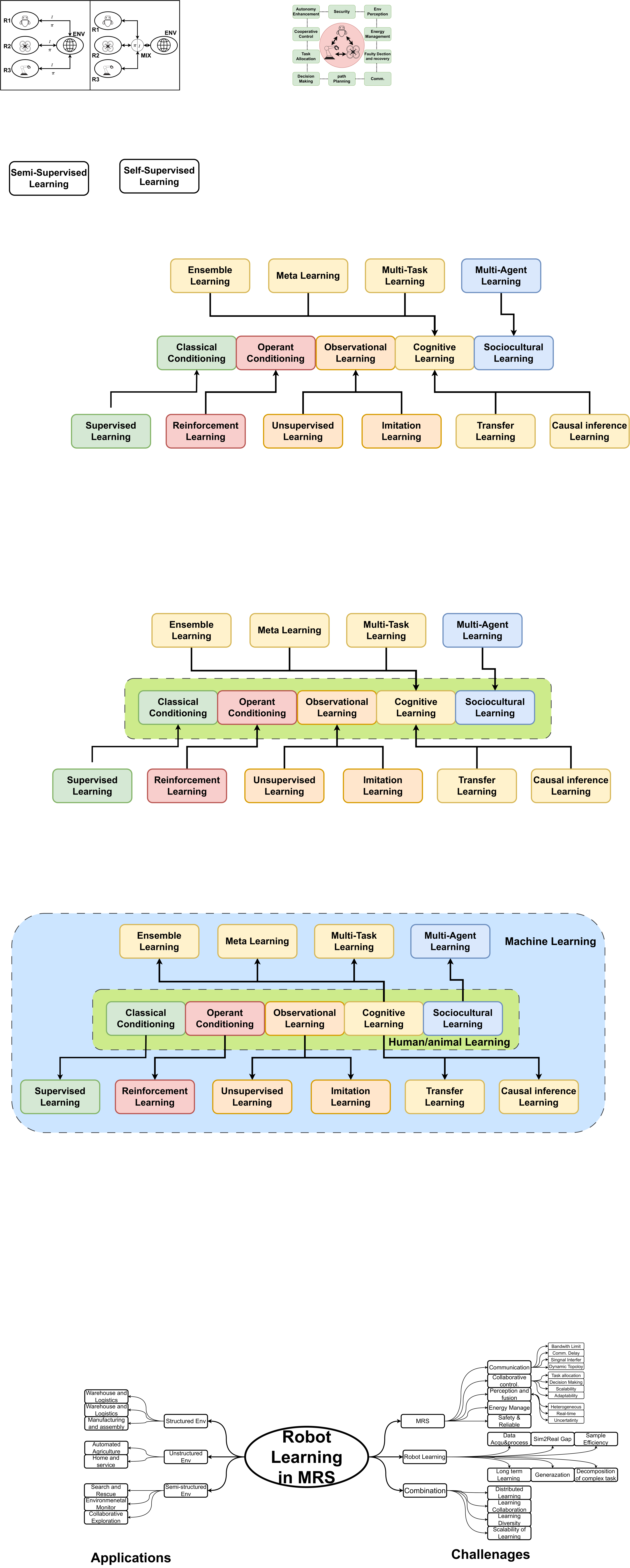}
\caption{Mapping of human/animal learning methods to machine learning methods}
\label{fig:learning method}
\end{figure*}

Although these do not cover all known ways of learning in human and animal, however, they do provide a brief classification basis for robot learning in MRS. Mapping the learning methods of human and animal to those of machine learning provides an interesting perspective on how robots emulate natural learning processes, as indicated in Figure \ref{fig:learning method}. The classification of learning methods based on carbon-based life forms such as human and animal can be mapped onto the principles of the learning processes of existing machine learning \(silicon-based\) methods.

\subsubsection{Classical Conditioning}
Classical conditioning involves learning through the association between stimuli. In the field of machine learning, the mechanism most similar to this associative learning is supervised learning \cite{bishop2006pattern, muhammad2015supervised}. This type of machine learning involves mapping between inputs (similar to conditioned stimuli) and outputs (similar to conditioned responses). Training data includes inputs and their corresponding outputs, and the model learns from these data to predict the output of new inputs. For example, when using neural networks for image recognition, the model learns the relationship between patterns in images and labels, similar to the associative learning between stimuli in classical conditioning. 

\subsubsection{Operant Conditioning}
Operant conditioning focuses on the relationship between behaviors and consequences, which is very similar to the concept of RL \cite{sutton2018reinforcement, kaelbling1996reinforcement}. In RL, an agent learns behavior strategies through interaction with the environment to maximize certain cumulative rewards. This learning process involves exploration (trying new behaviors to discover effective strategies) and exploitation (using known strategies to obtain rewards). The mechanism of RL is similar to operant conditioning, relying on the consequences of behaviors (rewards or punishments) to form or change behaviors.

\subsubsection{Observational Learning}
In machine learning methods analogous to observational learning, unsupervised learning \cite{bishop2006pattern, dike2018unsupervised} and imitation learning \cite{ravichandar2020recent, abbeel2004apprenticeship} stand out as particularly representative. Observational learning involves observing the behaviors and outcomes of others and learning from them. If one likens certain aspects of observational learning to feature learning or clustering in the field of machine learning, then the process in unsupervised learning, which involves identifying patterns and structures from data without explicit labels or feedback, shares strong similarities with observational learning. Inference learning (IL) is a method that allows robots or software agents to observe and mimic the behaviors of human experts or other agents. The key components of IL include: 1. Demonstrations - Behavioral demonstrations observed by the learner, usually provided by human experts or other advanced agents. 2. Behavior Cloning - A method of learning behavior acquired directly from demonstration, without the need for explicit modeling of the environment. 3. Inverse RL - Inferring a reward function through observation of demonstrations and then using this reward function to guide the learning process. It is evident that IL and observational learning of animals share a notable connection, with their principles being similar in many ways. Both learning methods involve learning from the observation of others' behaviors and imitating or replicating these observed behaviors in future actions.

\subsubsection{Cognitive Learning}
In machine learning, there are many methods that correspond to cognitive learning. For example, Transfer Learning (TL) \cite{pan2009survey, weiss2016survey} allows a model to apply existing knowledge (usually learned on one task) to another related but different task. This method is particularly useful in situations with limited data, as it can reduce the amount of data and computational resources needed to train on a new task. TL typically involves learning feature representations from a source task and then adapting these representations to a target task. 

Causal inference learning (CIL) \cite{yao2021survey, peters2017elements} focuses on learning the causal relationships between variables from data, rather than just correlations. This involves using statistical methods, experimental design, and computational models to determine whether one event (the cause) directly leads to another event (the effect) and how to quantify this impact. Both cognitive learning and CIL focus on understanding causal relationship. In cognitive learning, individuals understand causal relationships through observation, experience, and reasoning, while in CIL, algorithms identify and validate causal relationships through data analysis.

ML \cite{finn2017model} is defined as the process of "learning how to learn." The goal of ML is to enable machine learning models to optimize and improve the learning process through previous experiences, allowing them to adapt and learn more quickly when faced with new tasks. Both cognitive learning and ML involve higher-level learning on top of the basic learning process. In cognitive learning, this is manifested as metacognitive abilities \cite{salles2016metacognitive}, i.e., understanding and managing one's own learning process.  In ML, this is manifested as the ability to learn how to learn more effectively.

Ensemble learning (EL) \cite{dietterich2000ensemble, dong2020survey} improves the accuracy and stability of predictions by combining multiple learning models. The basic idea behind this method is that individual models may have their limitations, but when the predictions of multiple models (such as decision trees, neural networks, etc.) are combined, the overall predictive performance is enhanced through diversity and complementarity. In cognitive learning, individuals may use multiple sources of information and strategies to enhance learning effectiveness. EL improves predictive performance by combining multiple models. Both can potentially utilize a diversified perspective and methods to optimize results.

Multitask learning \cite{zhang2021survey, sener2018multi} enhances the model's generalization ability by learning multiple related tasks simultaneously during the training process. The core idea behind this method is that multiple tasks share some common representations or features so that while learning one task, the model can also learn useful information from other tasks. Both cognitive learning and multi-task learning may involve simultaneously dealing with multiple related tasks to improve efficiency and performance.

\subsubsection{Socialcultural Learning}
Sociocultural learning theory emphasizes that individuals learn and develop within the context of social interaction and cultural background. This type of learning naturally adapts to robot learning in MRS \cite{gronauer2022multi, anastassacos2021cooperation}, especially in decentralized systems where each robot can make independent decision. Multi-agent learning focuses on how multiple intelligent agents learn and interact in the same or interdependent environment. These agents learn the best strategies by observing the behavior of the environment and other agents to achieve their goals, which may involve cooperation or competition. Key issues in multi-agent learning include coordination, competition, communication, and the sharing of learning strategies. The similarity between sociocultural learning and multi-robot learning lies in the emphasis on interactions with other agents and the influence of the environment on learning.

\subsubsection{Affective Learning}
Affective learning in robots is a profound and complex research direction \cite{quick2022empathizing}, but this article will not delve into it extensively. However, when there is a human in the loop in MRS \cite{churamani2020continual, abendschein2021human}, the system may face issues related to recognizing, interpreting, processing, and simulating human emotions. These types of issues can be categorized under affective computing \cite{appriou2020modern}. How to properly handle this information to improve the naturalness and efficiency of human-computer interaction is a promising research direction.

\subsection{Reinforcement Learning}
The application of RL in MRS has been a research hotspot in the last decade. It is defined as a framework where each agent learns its behavior strategy through interaction with the environment to maximize certain cumulative rewards. Specifically, this learning process can be mathematically described as an extension of the Markov Decision Process (MDP), commonly known as the Multi-Agent Markov Decision Process (MAMDP) \cite{busoniu2008comprehensive, panait2005cooperative}.
A standard MAMDP can be defined as a tuple $(S, A, P, R, \gamma)$, where:
$S$ represents the state space, encompassing all possible environmental states.
$A=A^1\times A^2\times...\times A^n$ represents the joint action space, with each \(A^i\) being the action space for agent $i$.
$P: S \times A \times S \rightarrow [0,1]$ is the state transition probability function, indicating the probability of transitioning to the next state given the current state and joint actions.
$R: S \times A \times S \rightarrow \mathbb{R}$ is the reward function, which could be the sum of rewards for each agent or some other form of aggregation.
$\gamma$ is the discount factor used to calculate the present value of future rewards, with its value ranging from $0 \leq \gamma < 1$.
In a multi-agent environment, the goal of each agent is to learn a policy $\pi^i: S \rightarrow A^i$, aiming to maximize its expected cumulative discounted reward. Unlike single-agent RL, in multi-agent RL, each agent must consider the impact of the behaviors of other agents on the environment and on its own rewards.

For an agent, the objective can be mathematically defined as maximizing the expected cumulative discounted reward $V^\pi(s)$ or $Q^\pi(s, a)$, where $s$ represents the state and $a$ represents the action.
State-value function $V^\pi(s)$ represents the expected return under policy $\pi$ in state $s$. It is defined as the sum of expected rewards for all possible paths, where each reward is multiplied by the power of the discount factor $\gamma$, indicating the proximity in time:

\begin{equation}
    V^\pi(s) = \mathbb{E}_\pi\left[\sum_{t=0}^{\infty} \gamma^t R_{t+1} | S_0 = s\right]
\end{equation}

Action-value function \(Q^\pi(s, a)\) represents the expected return of taking action \(a\) in state \(s\), and then following policy \(\pi\):

\begin{equation}
    Q^\pi(s, a) = \mathbb{E}_\pi\left[\sum_{t=0}^{\infty} \gamma^t R_{t+1} | S_0 = s, A_0 = a\right]
\end{equation}

The goal of maximizing the reward function can be achieved by optimizing the aforementioned value functions. In a multi-agent environment, this goal is more complex because the optimal strategy of each agent may depend on the strategies of other agents. Therefore, agents need to learn a policy $\pi^*$ that maximizes their expected cumulative discounted reward while considering the possible strategies of other agents. For single-agent environment, this can be simplified to finding an optimal policy $\pi^*$ such that for all states $s$, $V^{\pi^*}(s) \geq V^\pi(s)$ for all $\pi$. Thus, the mathematical definition of maximizing rewards involves finding an optimal policy \(\pi^*\) such that:
For all states $s$, $V^{\pi^*}(s) = \max_\pi V^\pi(s)$, or for all states $s$ and actions $a$, $Q^{\pi^*}(s, a) = \max_\pi Q^\pi(s, a)$.
This is usually achieved through iterative algorithms such as dynamic programming, the Monte Carlo method, Temporal Difference learning, or deep learning, which gradually approximates the optimal policy $\pi^*$. 

%unique pros and cons
The basic idea of RL is to learn through interaction, which means that compared to traditional model-based methods, RL does not require detailed prior knowledge about the environment. Robots can learn effective strategies through trial and error, which is particularly useful in unknown or uncertain environments. For example, in \cite{patino2023learning, fan2020distributed}, RL is used to solve the navigation problems of robot teams in complex dynamic environment. At the same time, RL can improve the adaptability and flexibility of MRS, such as helping aerial robot swarms deal with complex turbulent flows 
 \cite{patino2023learning}, or MRS tracking of moving targets 
 \cite{guo2023cross}. RL can also handle multi-objective optimization problems, allowing each robot in an MRS to consider the overall system's optimal performance while pursuing individual goals. A continuous RL method introduced in 
 \cite{zhang2020continuous} can adapt to multi-objective optimization functions to guide robots' movement in dynamic environment. In the previous MAMDP definition, it is assumed that each robot is fully observable, but in reality, partially observable situations are more common. Fortunately, MAMDP can be easily extended to MAPOMDP, and in \cite{omidshafiei2017deep, foster2023complexity, hausknecht2015deep, wu2021impact}, all are based on the assumption of partial observability to apply RL to MRS. As a hot research topic in recent years, there are a rich set of references available for multi-robot reinforcement learning \cite{anastassacos2021cooperation, wang2023initial, nikkhoo2023pimbot, agrawal2023rtaw, fan2022welfare}.

On the other hand, RL also has unique disadvantages \cite{kapoor2018multi,foster2023complexity, nair2003taming,  zhang2021taming}. RL usually requires a large number of interaction samples to learn effective strategies, which may be impractical in real-world MRS due to the high cost and time consumption of physical experiments. For this reason, 
 \cite{zhao2020sim} discusses various aspects of the simulation to reality (Sim-to-Real) transfer problem in robotics Deep Reinforcement Learning (DRL). In \cite{wang2016sample}, an actor-critic algorithm combined with experience replay is introduced to improve sample utilization. By reusing past experiences, this method can learn effective strategies with fewer samples required. In complex multi-robot environment, ensuring the stability and convergence of learning algorithms is a challenge, especially in the case of continuous action spaces or multi-agent interactions. \cite{lillicrap2015continuous} introduces the Deep Deterministic Policy Gradient (DDPG) algorithm, a method that combines deep learning and reinforcement learning to solve control problems in continuous action space.

\subsection{Imitation Learning}
In the context of MRC, IL, as an effective learning strategy, aims to accelerate the learning process by observing and imitating the behavior of experts or other robots \cite{le2017coordinated,wang2023distributed, liu2023guide,ablett2023learning}.
While mathematical definition may vary depending on the specific method (such as Behavioral Cloning, Inverse RL, etc.) and the application scenario considered, a general mathematical framework can be stated: in the context of MRS where the followings are assumed, a state space $S$, an action space $A$, and a transition function $T: S \times A \rightarrow S$, representing the probability of system state transition under a given state and action. The goal is to learn a policy $\pi: S \rightarrow A$, that is, given the current state, to determine the action to be taken, in order to imitate the behavior of an expert (another robot or human). The expert's behavior is given by a set of trajectories $D = \{(s_1, a_1), (s_2, a_2), ..., (s_N, a_N)\}$, where $s_i$ represents the state and $a_i$ represents the action taken by the expert in that state. The goal of IL is to minimize the difference between the learning policy and the expert policy. This can be quantified in different ways, for example, by minimizing the distance between the policy output action and the expert action, or by maximizing the similarity of the trajectories generated by the learning policy to the expert trajectories.

IL has clear advantages and disadvantages in MRS. A unique advantage is that by observing and imitating robots or humans who have effectively performed specific tasks, other robots can quickly learn new skills, reducing the time needed to learn from scratch through trial-and-error. In \cite{liu2023guide}, a new active IL framework is proposed, where a teacher-student interaction model is utilized to significantly leverage the advantages of IL. Experiments on the MetaDrive benchmark and Atari 2600 games demonstrate that this method is more efficient in achieving performance close to that of experts compared to previous methods. 
In MRS, individuals can learn the importance of cooperation and cooperative strategies more quickly by observing the behavior of other cooperating robots, thereby promoting collaborative work throughout the group. In \cite{bara2022enabling} it is pointed out that the emergence of cooperative behavior can be explained through understanding the co-evolution process of cooperators' core and betrayers' periphery, emphasizing the role of partner selection and imitation strategies in promoting cooperative behaviors, without assuming the presence of underlying communication or reputation mechanism. In this way, the article provides a unified framework to study imitation-based cooperation in dynamic social networks.

The disadvantages of IL \cite{rajaraman2020toward} are also obvious.  They can be summarized into the followings:
1. Dependence on high-quality demonstrations: If the quality of expert demonstrations is not high, or if they do not match the current task environment, the learned strategy may lead to poor performance or even incorrect behavior.
2. Limitations in generalization capability: Since IL relies on specific demonstrations, the learned strategy may exhibit insufficient generalization capability when encountering unseen environment or task.
3. Lack of self-exploration: Over-reliance on imitation may lead to a lack of self-exploration and innovation capability in robots, preventing them from autonomously discovering solutions that are superior to the demonstrations.
4. Lack of diversity: In an MRS, if all robots imitate the same demonstration, it may lead to homogenization of behaviors, reducing the system's adaptability and robustness. 
With the development of IL, methods to address flawed demonstrations have been proposed in \cite{li2024imitation,kim2021demodice}. A novel IL framework introduced in \cite{liu2024ceil} expands the applicability of IL by incorporating the concepts of hindsight information embedding and contextual strategies, demonstrating superior performance across multiple tasks and settings.

\subsection{Transfer Learning}
In the context of MRS, the concept of TL typically involves transferring knowledge learned on one robot or a group of robots to other robots or robot groups, in order to improve learning efficiency, reduce the amount of training data needed, or enhance performance in new tasks \cite{schwab2018zero, chen2018hardware, pereida2018data, chen2023transfer,smith2023learning}. 
While there are detailed mathematical models and definitions for specific applications, a general mathematical framework for TL may involve the following key elements: 
1. Source Task $T_S$ and Target Task $T_T$, which are defined through task-related data distributions, objective functions, etc.
2. Source Domain $D_S$ and Target Domain $D_T$, each domain consists of a feature space and a marginal probability distribution (i.e., $P(X)$). In MRS, different robots may encounter different environment (domain).
3. Transfer Function $f$, which is the mapping from the source task to the target task. The purpose of this function is to enable the effective application of the knowledge learned on the source task to the target task. \cite{helwa2017multi} introduces a framework for multi-robot TL from a dynamical system perspective.

From the perspective of multi-robot TL, TL can assist in the learning process of robots, reducing the training time and data needed to achieve excellent performance, while also promoting the sharing of knowledge and experience. This enables robots to learn not only from their own experiences but also from the successes and failures of other robots which is similar to IL. Especially under resource-constrained conditions (such as computing power, storage space, etc.), TL can reuse existing knowledge. As  for shortcomings, if the source task and target task are not sufficiently similar, or if the method of TL is not properly implemented, negative transfer may occur \cite{vrbanvcivc2020transfer,gui2018negative}. This means the knowledge learned from other robots could actually decrease the performance of the robot on the target task. Moreover, to achieve effective knowledge transfer, communication and coordination among robots are necessary, which might increase the complexity and overhead of the system.

\subsection{Causal Inference Learning}
Causal inference learning refers to the process of using data to infer the causal relationships between variables. In the context of MRS, causal inference can help comprehend and predict the interactions between different robot behaviors and environmental factors \cite{luo2020causal,wang2023deconfounded, guo2023enabling, smith2018distributed}.
Although the concept of causal inference has a precise mathematical definition in statistics, its application in MRS remains an active research area, involving complex dynamic systems and interactions. In causal inference research, a core concept is the Potential Outcomes Framework, also known as the Rubin Causal Model (RCM). Additionally, methods based on Graphical Models play an important role in causal inference, especially in describing complex causal relationships between variables. Under the Potential Outcomes Framework, for each individual and each possible treatment (or intervention), there is a potential outcome. Causal effect is defined as the difference in potential outcomes under different interventions. Mathematically, if $Y_i(t)$ represents the potential outcome of individual $i$ under intervention $t$, then the causal effect for individual $i$ can be expressed as $Y_i(t_1) - Y_i(t_0)$, where $t_1$ and $t_0$ represent different intervention states. Graphical models use graphs (typically Directed Acyclic Graphs, DAGs) to represent the causal relationships between variables. In these models, nodes represent variables, and directed edges represent causal relationships. Through the analysis of the graph, one can identify conditional independency, causal pathway, and possible intervention effect. In MRS, CIL usually focuses on how to infer the causal relationships between robot behaviors based on observed data (e.g., the behavior of robots and changes in the environment) or the causal effects of robot behaviors on environmental changes. This includes understanding how the behavior of one robot might affect the behavior of other robots or the overall state of the system.

\begin{figure}[htbp]
\vspace{4pt}
\centerline{\includegraphics[width=0.4\textwidth]{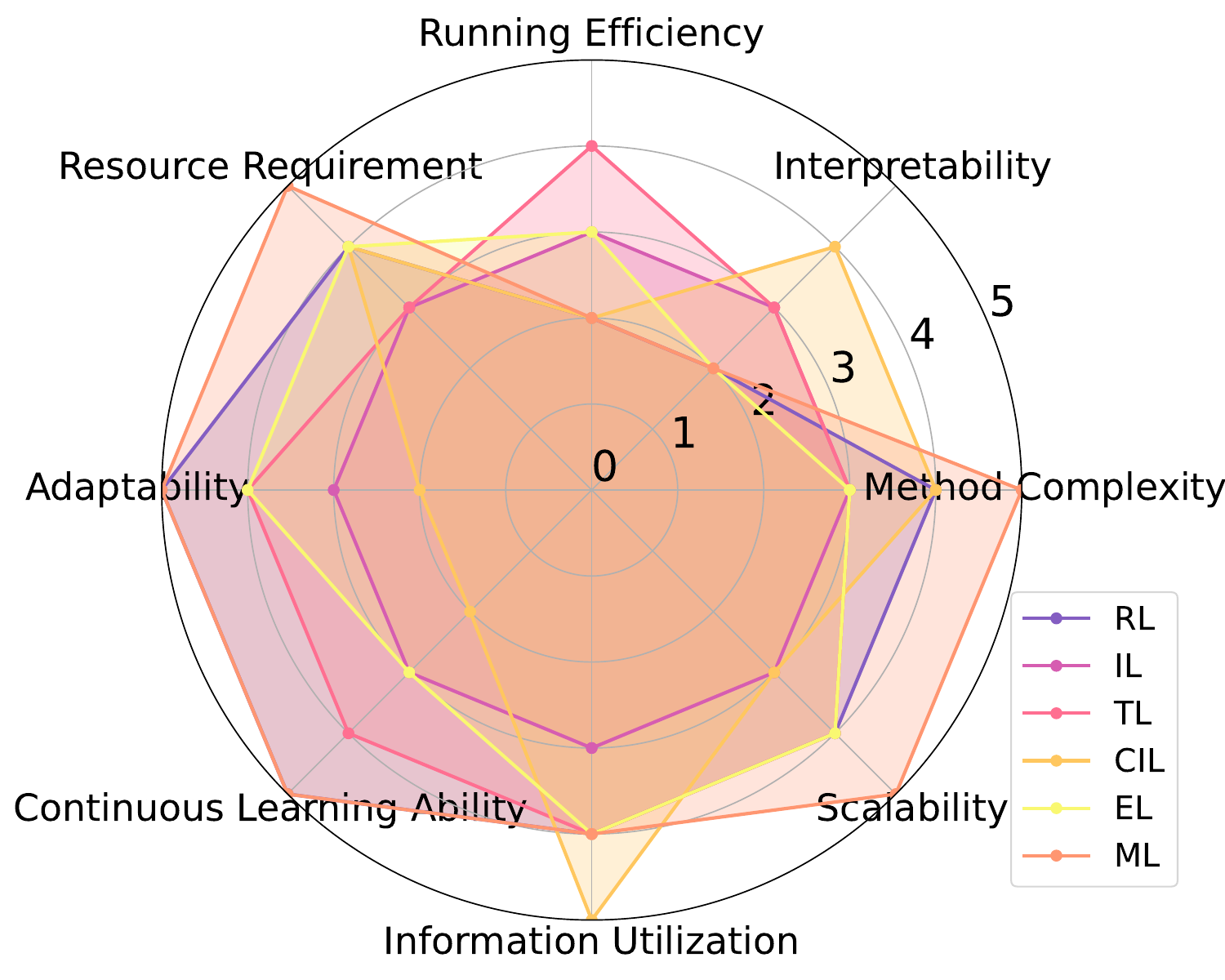}}
\captionsetup{font=footnotesize}

\caption{Evaluation of six robot learning methods across eight dimensions}
\label{fig: rader}
\end{figure}

\begin{table*}[!t]
\renewcommand{\arraystretch}{1.3}
\caption{Analysis of literature on robot learning methods used in MRSs in the article}
\label{publication summary}
\centering
\begin{tabular}{lp{1.7cm}p{2cm}p{2cm}p{5cm}p{4cm}}
\hline
Ref & Methodology & Application & Topology & Joint Decision making & Limitations \\
\hline
\cite{panait2005cooperative} & Q-Learning & Non-specific & Decentralized & Based on the reputations of others, derived from past interactions. & May converge to non-efficient outcomes \\
\cite{omidshafiei2017deep} & DRQNs & Underwater vehicles & Decentralized & Distillation of single-task policies into a unified policy that performs well across multiple tasks, without explicit task identification during operation. & Complexity of training \\
\cite{wang2023initial} & Attention-Based DRL & Surveillance & Hybrid & By the DRL model & Complexity of training \\
\cite{guo2023cross} & CE-PG & Search and Rescue & Hybrid & Learned policy and their current state & Complexity of implementing the scheme \\
\cite{nikkhoo2023pimbot} & PG & Non-specific & Hybrid & Manipulated incentives and policies designed & Complexity of implementing the scheme \\
\cite{foster2023complexity} & MA-DEC & Non-specific & De/Centralized & Structured observation framework & Scalability, complexity of implementing the scheme \\
\cite{patino2023learning} & GCNN, DRL & Aerial Operation & Decentralized & Sharing sensor measurements between nearby robots & Sim2Real gap \\
\cite{kurzer2022learning} & IRL, MCTS & Automated vehicles & Decentralized & Mimic human-like behavior in traffic & Complexity in learning effectively \\
\cite{wang2023scrimp} & IRL, transformer & Non-specific & Decentralized & Local observations, global communication inputs, and a learning-based stochastic tie-breaking strategy & Complexity in implementation scheme, scalability \\
\cite{bara2022enabling} & EGT & Non-specific & Decentralized & Imitating the strategies of their neighbors in the network & Complexity of implementing the scheme \\
\cite{helwa2017multi} & ODM & Quadrotor & De/Centralized & The emphasis is on the transfer learning process between individual robots rather than collective decision-making. & Poor generalization ability \\
\cite{chen2018hardware} & DRL, HCP & Manipulation and Locomotion & De/Centralized & Optimizing individual policies based on the hardware characteristics of each robot. & Poor generalization ability \\
\cite{pereida2018data} & AC ILC & Trajectory Tracking & Centralized & Transferring learned control inputs for accurate trajectory tracking & Poor generalization ability \\
\cite{venturini2020distributed} & DDQL & Monitoring and Surveillance & Decentralizedshared & Experiences and a common learning coordinating implicitly through the distributed reinforcement learning framework to achieve collective objectives. & complexity of computing, scalability \\
\cite{guo2023enabling} & PM & Search and Detection & Hybrid & Trust levels are made jointly through the aggregation of direct and indirect experiences. & Poor in limited communication \\ 
\cite{wang2023deconfounded} & CI, RL & Football Strategy & Decentralized & Causal relationships between players and opponents & Sim2Real gap, complexity of implementing the scheme \\ 
\cite{smith2018distributed} & DIPM & Search and Rescue & Decentralized & Conflicting goals are resolved through a single-bid auction mechanism among locally communicating robots & Complexity of implementing the scheme \\ 
\cite{wu2022evaluating} & MARL, RL & Non-specific & Decentralized & Reinforcement learning policies that evolve from individual experiences. & Complexity of implementing the scheme \\ 
\cite{lu2023preference} & DWPI & Non-specific & Decentralized & Inferring other agents' preferences & Poor generalization ability \\ 
\cite{ghosh2021deep} & DE-DRL & Air Traffic Control & Decentralized & Arbitrating between the decisions made by the local kernel-based RL model and the wider-reaching deep RL model, leveraging the strengths of both methods. & Complexity of implementing the scheme and training \\ 
\cite{kotary2022end} & e2e-CEL & Non-specific & De/Centralized & By learning to select and aggregate predictions from a subset of base learners. & Complexity of implementing the scheme and training \\ 
\cite{lou2023pecan} & CA-PE & Overcooked Environment & Decentralized & Context-aware module that predicts the partner's policy level, & Complexity of implementing the scheme and training \\ 
\cite{edwards2023collaborative} & NMEM & Monitoring & Decen/Hybrid & Combination of local collaboration signals and a macroscopic model & Scalability \\
\cite{zhao2023stackelberg} & SML & UAVs & Decentralized & Dynamic Stackelberg game & Complexity of implementing the scheme \\ 
\cite{yang2021adaptive} & MGRL & power grids, traffic tolling systems & Hybrid & Individual reward functions modified by incentives from the central planner, & Sim2Real gap, high computing source \\ 
\cite{yel2022meta} & ML & UAVs & Centralized & Enables an individual UAV to autonomously detect unsafe situations and replan its trajectory. & Sim2Real gap, Complexity of implementing the scheme \\ 
\cite{kayaalp2022dif} & Dif-MAML & Monitoring & Decentralized & Achieve consensus on a common "launch model" & Poor in limited communication \\

\hline
\end{tabular}
\end{table*}

MRS typically involve multiple autonomous robots interacting in a shared environment to complete various tasks, such as collaborative transport, search and rescue, and automated monitoring.  CIL has a unique advantage in explainable robot learning methods, as it can help one understand how the behavior of one robot affects the behavior of other robots or the overall state of the system. This is crucial for designing highly coordinated and efficient MRS. Furthermore, by identifying and understanding causal relationships, system designer can better formulate intervention measures (such as adjusting task assignments, communication strategies, etc.) to optimize the performance of the entire system. When robots clarify the various causal structures between themselves and the environment, they can greatly reduce the dependency on large-scale data, which is significant in situation where data is scarce or the cost of data acquisition is high.
On the other hand, the dynamism and interaction complexity of MRS make constructing accurate causal models very challenging. Therefore, the identification and verification of causal relationships require precise model design and complex data analysis \cite{wu2022evaluating,xu2023causal}. Although causal inference can reduce the dependency on large amounts of data, it still requires high-quality data to identify true causal relationships. Collecting and organizing such data in MRS is also very challenging. Conducting experiments in MRS (such as randomized controlled trials) to verify causal relationships may be impractical, especially in real-world applications, such as operating in unstable or uncontrollable environment  \cite{lu2023preference, wang2013probabilistic}.

\subsection{Ensemble Learning}
In the context of MRC, Ensemble Learning (EL) can be seen as multiple robots (or agents) working together to improve the effect of overall task execution, where each robot can be considered as a base learner. 
The mathematical definition of EL usually relates to a specific ensemble method, such as Bagging, Boosting, or Stacking. Generally, the goal of EL is to combine the predictions of multiple models to reduce generalization error. Mathematically, EL can be defined as suppose there is a set of base learners $\{h_1, h_2, ..., h_T\}$, each learner $h_i$ can give an output $h_i(x)$ for a given input $x$. The goal of EL is to combine these predictions through a certain strategy (such as simple averaging, weighted averaging, or voting) to form the final ensemble prediction $H(x)$:

\begin{equation}
    H(x) = f(h_1(x), h_2(x), ..., h_T(x))
\end{equation}

\noindent where $f$ is the combining strategy, and $T$ is the total number of base learners. In the context of MRC, the decision or prediction of each robot can be considered as the output of a base learner, and the ensemble method is used to coordinate the decisions among these robots, in order to improve the overall efficiency or accuracy of task execution.

In MRS, the greatest advantage of EL stems from its fundamental characteristic of improving prediction accuracy, robustness, and generalization ability by combining multiple models. By adjusting EL strategies, MRS can flexibly adapt to changes in task requirements or robot capabilities \cite{lou2023pecan, ghosh2021deep, yu2019decentralized, kotary2022end, edwards2023collaborative}. Supriyo Ghosh \cite{ghosh2021deep} combined the strengths of both kernel-based and deep multi-agent RL policies. This combination allows the system to leverage fine-grained local policies and more global policies efficiently, improving the decision-making process in air traffic control scenarios. \cite{yu2019decentralized} presents a decentralized EL approach that leverages sample exchange among multiple agents to improve performance in multi-agent systems. The method allows for decentralized data handling by having agents exchange data samples to enhance their collective predictive abilities. The benefits of using this ensemble method include increased accuracy through collaborative learning, competitive performance with state-of-the-art methods while maintaining data decentralization, and efficient utilization of local data resources by each agent. \cite{kotary2022end} introduces a novel framework for EL through differentiable model selection, integrating machine learning with combinatorial optimization.  Despite the many advantages that EL offers in MRS, realizing these advantages also requires addressing a series of challenges, including how to effectively integrate data and decisions from different robots \cite{rincy2020ensemble,mohammed2023comprehensive}.

\begin{figure*}[htbp]
\centering
\vspace*{5pt}
% 第一个subplot
\begin{subfigure}[b]{0.45\textwidth} % 调整宽度以适合你的需求
    \centering
    \includegraphics[width=\textwidth]{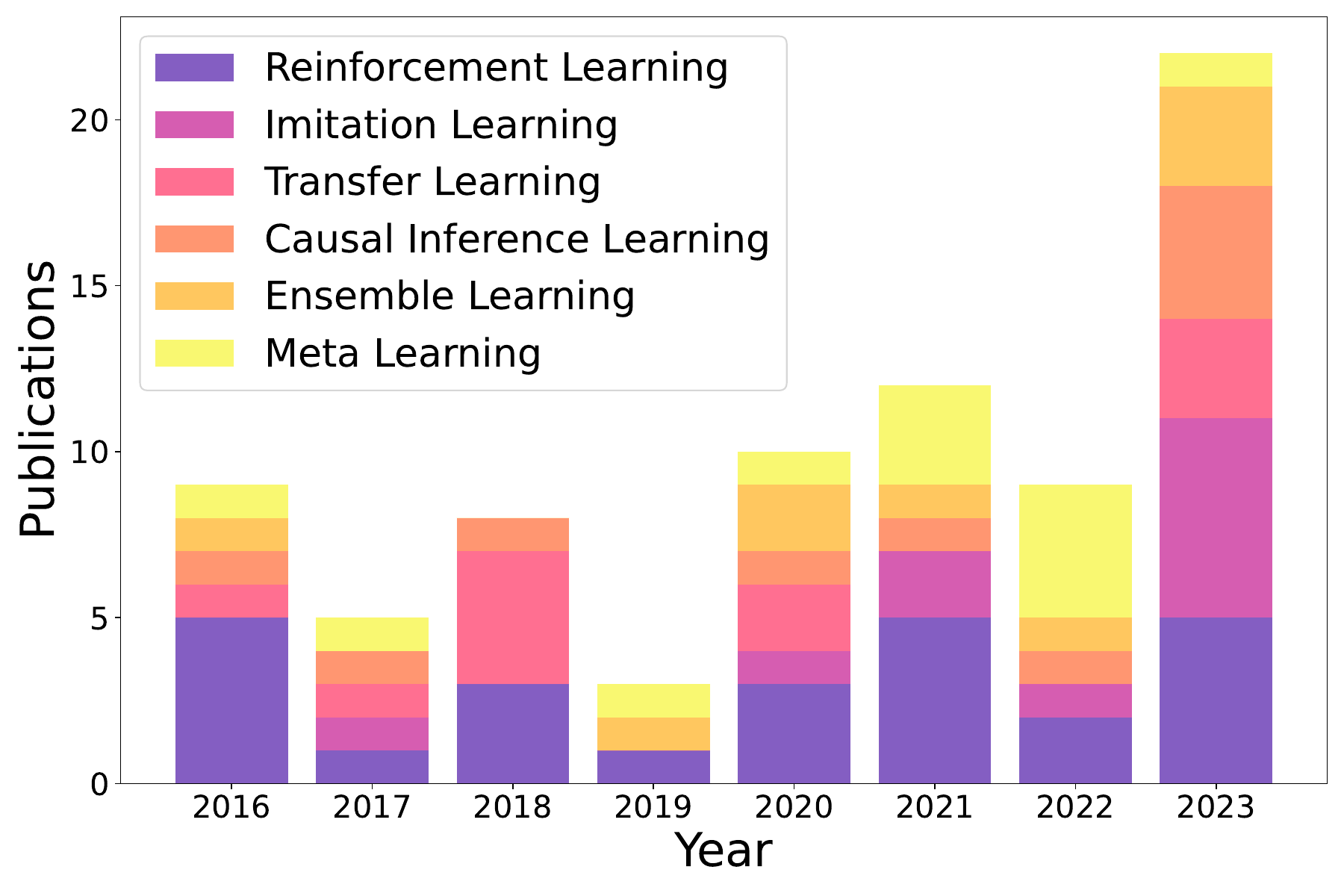} % 修改为你的图片文件名
    \caption{}
    \label{fig:year stack}
\end{subfigure}
\hfill % 添加一些水平空间
% 第二个subplot
\begin{subfigure}[b]{0.45\textwidth} % 调整宽度以适合你的需求
    \centering
    \includegraphics[width=\textwidth]{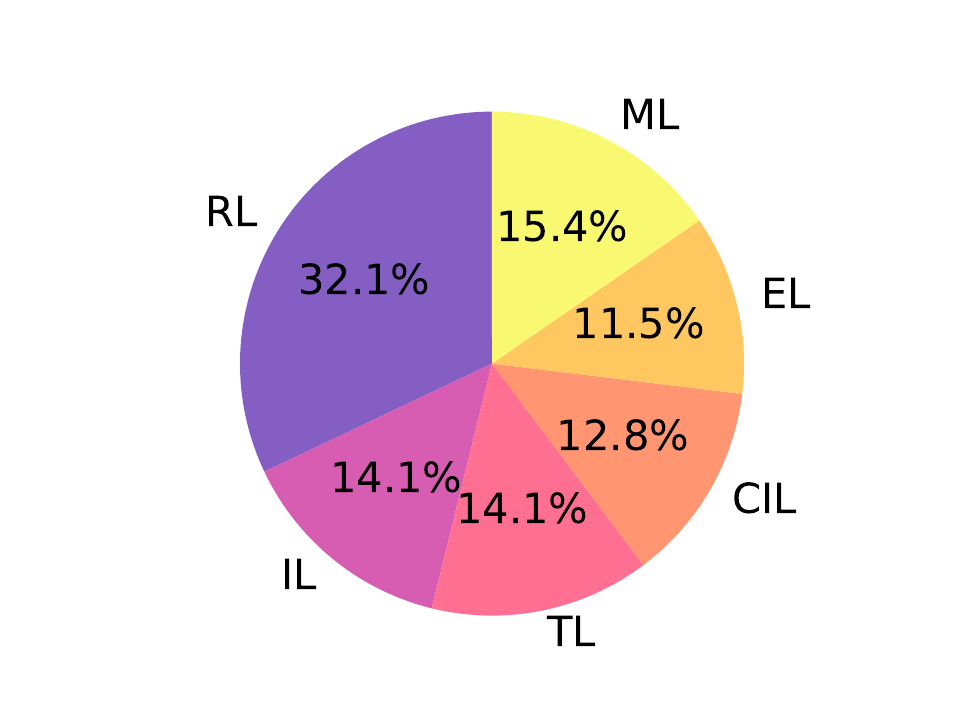} % 修改为你的图片文件名
    \caption{}
    \label{fig:method category pie}
\end{subfigure}
\caption{Statistics on the number of cited papers: a. citation counts of different robot learning method articles by year; b. Proportion of different robot learning methods }
\label{fig:year}
\end{figure*}

\subsection{Meta Learning}
Meta learning (ML), in the field of machine learning, refers to the process of building models that learn how to learn. It enables models to use past experiences to accelerate the learning process for new tasks, or to improve performance on new tasks \cite{finn2019online,zhao2023stackelberg, yang2021adaptive, schrum2022meta, yel2022meta}. 
Although the concept of ML is relatively intuitive, its mathematical definition may vary depending on different research background and application scenario. In the context of MRC, the goal of ML is to enable robots to quickly adapt to new cooperative tasks or environments drawing from previous cooperative experiences. In this scenario, ML often focuses on learning how to effectively share information, make decisions, and adapt to the behaviors of cooperative partners. A summary of all the methods mentioned above and their corresponding references can be found in Table \ref{publication summary}.

Mathematically, ML in MRS can be defined as finding a learning algorithm $\mathcal{A}$, which use the learning experiences from past tasks $\{T_1, T_2, ..., T_n\}$ to improve learning efficiency and performance in a new task $T_{new}$. Specifically, consider a system containing multiple robots, each robot $i$ has a parameter vector $\theta_i$ for task $T$, and the system's goal is to minimize a common loss function $L(T, \{\theta_i\})$, which measures the performance of the entire robot team in task $T$.
The process of ML can be seen as finding an optimization algorithm $\mathcal{A}$, which adjusts the learning strategy of each robot so that the team can adapt more quickly to new tasks. This can be formalized by minimizing the expected loss over all tasks, as below:

\begin{equation}
\min_{\mathcal{A}} \mathbb{E}_{T \sim \mathcal{T}}[L(T, \{\theta_i^{*}\})]
\end{equation}

\noindent where \(\{\theta_i^{*}\}\) is the set of robot parameter vectors obtained by applying algorithm \(\mathcal{A}\) to task \(T\), and \(\mathcal{T}\) is the distribution of tasks.

Applying ML in MRS offers a series of unique advantages. For instance, MRS can quickly adjust their strategies to adapt to new environments or tasks through ML, reducing the exploration time in unknown environments \cite{mohammadi2020introduction, hu2021distributed,kayaalp2022dif}. Moreover, once a single robot learns a new skill or strategy, this knowledge can be rapidly disseminated throughout the entire robot group via ML mechanisms, improving the overall learning efficiency. This also facilitates the sharing of strategy and experience among robots, enabling the entire system to perform complex tasks cohesively. In summary, ML allows robots to learn from past experiences to adapt quickly to new task or environment, which is especially important for MRS, as these systems often need to operate collaboratively in dynamically changing environment. \cite{saunshi2020sample, gupta2021dynamic}
%合作策略：分析多机器人合作中的不同策略和方法。
%案例研究：提供一些成功应用机器人学习进行多机器人合作的实例。
%效果评估：探讨这些方法在实际应用中的效果和表现。

\section{Discussion}
\subsection{Technical Challenges}
%先介绍以下robot learning在多机器人系统中遇到的具有共性的技术挑战，在通讯，共同决策上面临的挑战，在说明一下各个机器学习算法如何面对这些挑战
Technical challenges of robot learning in the context of MRS mainly come from three aspects: 1. The complexity of MRS themselves \cite{rizk2019cooperative, rogers2012delivering, foster2023complexity,yan2013survey}, 2. The intrinsic complexity of specific robot learning methods \cite{kroemer2021review, rajaraman2020toward}, and 3. Potential new problems that arise from the complexity resulted from merging the first two.  \cite{kapoor2018multi,nair2003taming, zhang2021taming}. 

MRS face numerous challenges, especially in complex environments such as underground, underwater, or remote areas, where communications can be severely limited \cite{gautam2012review, doriya2015brief}. This includes issues like communication delay, signal attenuation, and data loss, which pose challenges to real-time coordination and control. Moreover, effectively allocating tasks and coordinating control to optimize use of resource, improve efficiency of task execution, and adapt to dynamically changing environments remains a difficult problem \cite{smith2018distributed, khamis2015multi}. In terms of perception \cite{queralta2020collaborative}, compared to SRS, MRS face more difficult challenges in dealing with the heterogeneity, uncertainty, and temporality of different sensor data, as well as the challenge of processing large volumes of data in real-time with limited computing resources. In certain complex environments, robots in an MRS need to understand the intentions and behaviors of other robots to work effectively together. This requires not only advanced communication capability but also the ability to engage in complex social interaction and collaborative decision-making. A lot of works remain required to address other issues such as safety, reliability, and scalability still.

In the previous section on learning methods, the respective weakness and challenge of each robot learning method were discussed. Challenges that are common across these robot learning methods \cite{nair2003taming, wang2016sample} are considered in the present section. First is data and sample efficiency. Robot learning is often limited by the amount of available training data. Collecting a large volume of labeled data in the real-world is both expensive and time-consuming. Therefore, improving learning algorithms' data efficiency, such as through transfer learning, learning from simulation, and sample-efficient strategy in RL, is a significant challenge. Moreover, the generalization ability of learning methods is also an essential issue \cite{liu2024ceil,cobbe2019quantifying}. Trained robot models need to perform well in unseen environment and situation. Enhancing generalization ability requires an algorithm not only to learn patterns in the training environment but also to adapt to new and dynamically changing environment. Real-time decision-making and control are essential \cite{bajcsy2019scalable, derbas2014multi}, as robots must be able to respond quickly to environmental changes and make decision and take action. This requires learning algorithms to achieve efficient data processing and decision-making with limited computing resources.

\begin{figure}[htbp]
\vspace{4pt}
\centerline{\includegraphics[width=0.4\textwidth]{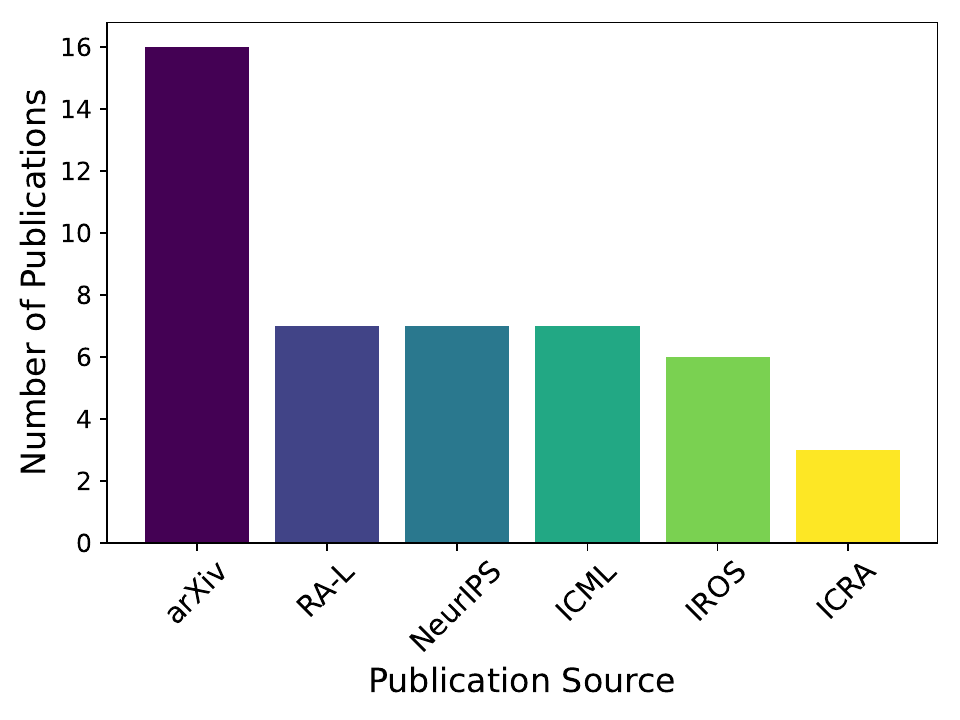}}
\captionsetup{font=footnotesize}
\caption{Top six sources of cited literature}
\label{fig: publication}
\end{figure}

Lastly, whether these learning methods can autonomously learn and adapt is critical \cite{churamani2020continual, parisi2019continual, shaheen2022continual}. In complex and constantly changing environment, robots need to have the ability to learn and self-optimize in real-time. This demands the development of learning mechanisms that can automatically adapt to new task and environment without direct human intervention or disruption.

\subsection{Applications}
Applications of Robot Learning in MRS can be categorized based on the known extent of the environment into three types: structured environment, semi-structured environment, and unstructured environment, see figure\ref{fig:application and challenges}. This classification helps in understanding the challenges and requirements of robot learning and collaboration in different settings.

\begin{figure}[htbp]
\vspace{4pt}
\centerline{\includegraphics[width=0.4\textwidth]{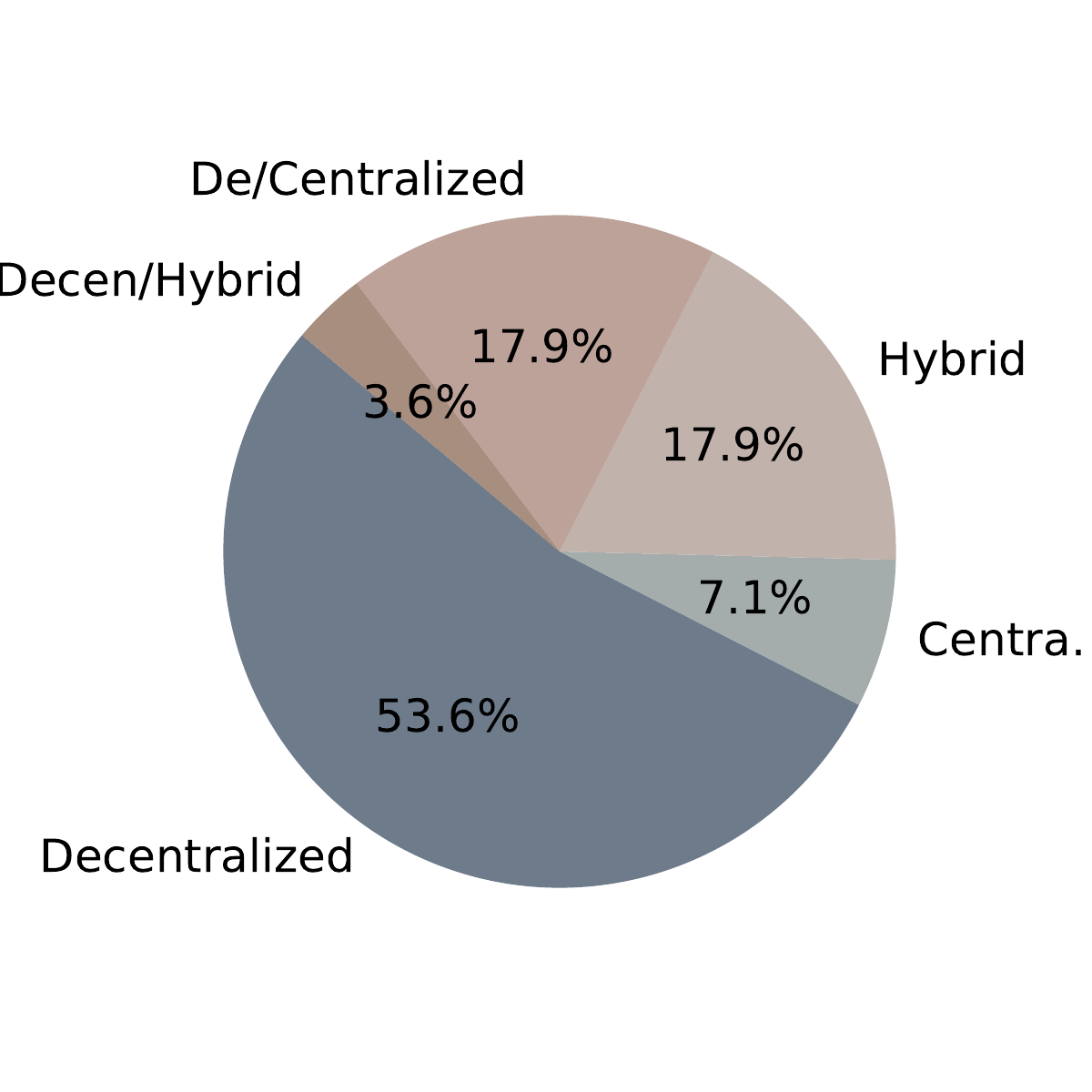}}
\captionsetup{font=footnotesize}

\caption{Proportion of different communication topologies}
\label{fig: topology pie}
\end{figure}

In structured environment, layout and condition are relatively invariant and predictable. Robots can rely on accurate maps and predefined rules for operations. For example, in warehouse logistics such as cargo handling, sorting, and distribution \cite{agrawal2023rtaw}, the environments are relatively unchanging, with shelf and aisle layouts being predetermined and known. On manufacturing lines \cite{chen2020wireless}, where tasks and operational procedures are standardized, robots can learn specific assembly skill and collaboration pattern.
Unstructured environments are highly dynamic, unpredictable, and lacking set rules and structures, demanding higher autonomy and learning capability from robots. For instance, in the aftermath of earthquake or flood where the environmental condition is highly complex and uncertain, robots would be required to autonomously explore and carry out rescue missions \cite{queralta2020collaborative,nazarova2020application}. Similarly, monitoring in natural settings such as forest, ocean, or polar region involves complex and variable environmental conditions. Exploring unknown or hazardous environments like deep sea, underground cave, or outer space, where the environment is entirely unknown, necessitates autonomous learning and collaboration by robots \cite{olcay2020collective, bai2020cooperative}.
Semi-structured environment lie between structured and unstructured environments, featuring some predictability but also variability and uncertainty, requiring robots to have a degree of adaptability and learning capability. A good example is automated agriculture \cite{lytridis2021overview}, where the basic layout of a farmland is a constant, but factors such as crop growth condition, weed emergence, and weather condition introduce variability. Similarly, domestic and household service robots have to deal with a constant basic layout but face significant variability in the placement of daily items and the activities of family members \cite{abou2020systematic}.
Through the application of MRS in real-world scenarios, the specific requirements and challenges of Robot Learning in different environments must be defined and available to enable targeted design and optimization of learning strategies and collaboration mechanisms.

In the context of MRC, robot learning faces a series of unique challenges, reflecting issues from different dimensions such as distributed learning, collaborative learning, diversity in learning, to scalability of learning. The following is a discussion about these four aspects of challenges:
In distributed learning \cite{fan2020distributed, smith2018distributed, choi2009distributed}, each robot may operate in different environments, collecting data with different characteristics and distribution needs. How to handle this data heterogeneity to achieve effective distributed learning is a problem.
In collaborative learning \cite{edwards2023collaborative, queralta2020collaborative}, robots in an MRS need to coordinate their actions to achieve a common goal. How to design learning algorithms to discover and implement efficient collaboration strategies to adapt to complex tasks and environments is a key challenge.
In learning diversity \cite{yang2021adaptive}, robot systems need to be able to adapt to dynamic changes in the environment. This requires learning algorithms to handle the uncertainty and variability of the environment, while maintaining adaptability to diversity.
In terms of algorithm and system scalability \cite{bajcsy2019scalable, rana2000scalability, hsu2021scalable}, as the number of robots increases, how to maintain the scalability of learning algorithms and systems becomes a major challenge. Algorithms need to be able to efficiently process the data generated by a large number of robots, and the system design should support the addition of more robots without degrading performance.

\subsection{Quantitative Analysis}
In Figure \ref{fig: rader}, eight dimensions are used to guage the applicability of various robot learning methods in the context of MRC: running efficiency, interpretability, method complexity, scalability, information utilization, continuous learning ability, adaptability, and resource requirement. Each dimension is rated on a scale of 1-to-5 to indicate its strength or weakness. RL and ML perform well in six of these dimensions while showing shortcomings in running efficiency and interpretability. Albeit having a balanced evaluation across all the eight dimensions, nevertheless, IL fails to excel in any of the specific areas. CIL has the best interpretability and information utilization among all the methods, however, it performs worst in running efficiency. TL's adaptability falls between RL and IL. EL shows deficiencies in interpretability and continuous learning ability compared to other dimensions. 

In Figure \ref{fig:year stack}, the statistics on the year of publication of the articles cited in the paper and the corresponding number of robot learning articles are compiled. It's important to note that the statistics for 2016 include both 2016 and the preceding years, and the statistics for 2023 include both 2023 and 2024. From the figure, it's evident that focus is given to articles published in the last seven years, especially those published in 2020 and later, to better illustrate current research progress. In the pie chart seen in Figure \ref{fig:method category pie}, the proportion of articles on the six most discussed robot learning methods in this text is calculated. It is evident that articles on RL register the largest number, accounting for 32.1\% of the total, while the least is ensemble learning, which accounts for 11.5\%. The numbers for other articles are relatively close. In Figure 
 \ref{fig: publication}, the top six sources of the most applied articles in this text are listed. Arxiv preprints and conference papers dominate due to their quick publication cycles, which is closely related to the faster pace of research progress in robotics and machine learning compared to other fields in recent years. Figure \ref{fig: topology pie} shows the proportion of communication topology structures in robot learning methods for MRS reviewed in the previous sections. Methods that are decentralized account for more than 50\%, while hybrid methods and those supporting both decentralized and centralized approaches each account for 17.9\%, with centralized methods being the least, at only 7.1\%. This aligns with one of the assumptions mentioned earlier that individuals in an MRS can make decisions independently. In summary, the proportion of the articles cited in this paper exploring decentralized methods reaches 92.9\%.
\subsection{Research Trending}
%学习算法和mrc的结合可以带来哪些变革，例如，明确性环境，仓库，封闭环境，不确定环境，灾后救援，受限环境，水下无信号
%前文中没有提及的一个新的方法例如大语言模型，和一些趋势
%是否中心化， 是否通讯受限，是否。。。
Robot learning in the context of MRC is facing a trend of rapid development and ongoing evolution. Existing methods have already achieved good results in many customized tasks, but there are still many challenges in generalized task requirement and under restricted condition. While predicting the future is difficult, nevertheless, a few key driving factors for robot learning in MRS from a realistic perspective are considered in the followings. Considering the balance between generality and customization, large language models (such as the GPT series) offer powerful natural language processing capabilities, which can facilitate robots in understanding and executing more complex instructions. In the future, one can foresee these models being further customized to fit specific MRC scenarios while maintaining a degree of generality to flexibly handle various tasks. Moreover, as the capability of large language models in understanding and generating natural language continues to improve, interactions between robots as well as between robots and humans will become more natural and efficient. This will greatly enhance the collaborative efficiency of MRS in executing complex tasks. In practical complex tasks, a single robot learning method may not be sufficient to meet requirements, and combining different machine learning techniques can provide a more flexible and powerful learning mechanism. Especially in future's broader human-robot interactions, the greatly increased interpretability of robot learning brought by causal inference can help both robot systems and humans understand the causal relationships behind tasks more deeply, thereby making more reasonable and efficient joint decisions. As robot technology develops, how to effectively reduce energy consumption becomes an important topic. Optimizing algorithms and hardware design, as well as adopting more efficient learning methods, will be key to reducing the energy consumption of MRS.

\section{REMARKS}
This article performed a comprehensive survey on the wide range of mainstream methods and frameworks of robot learning within the context of MRS. Instead of directly discussing the classification and differences of learning methods from the perspective of machine learning, human and animal learning methods was first mapped to robot learning approaches. Especially in the current era of rapidly expanding intelligent robotics, the presented review provides a summary and blueprint for future research in robot learning, particularly in the context requiring collaborative learning. Moreover, considering the current trend of technology and development, robot systems are expected to be more widely applied in all aspects of human life. The frequency of interactions between robots and humans, animals, other robots, and the entire environment will increase. Promoting these interactions, robot learning technology is currently an important research topics with an implication to the foreseeable future.

\end{document}